\documentclass[english, letterpaper, 10pt]{ieeeconf}

\makeatletter
\IEEEtriggercmd{\reset@font\normalfont\scriptsize}
\makeatother
\IEEEtriggeratref{1}

\usepackage{subcaption}
\usepackage{multicol}
\usepackage{amssymb}
\DeclareCaptionLabelSeparator{periodspace}{.\quad}
\usepackage[dvipsnames]{xcolor}
\usepackage[normalem]{ulem}

\usepackage[ruled,noend,linesnumbered]{algorithm2e}
\usepackage{float}

\usepackage{xr-hyper}

\SetKwInput{KwInput}{Input}                
\SetKwInput{KwOutput}{Output}              

\usepackage{algpseudocode}

\usepackage{tikz}
\usepackage{textcomp}
\usepackage{lipsum}
\usepackage{footnote}  
\usepackage{booktabs}

\usepackage{array}
\usepackage{makecell, caption}

\usepackage{bbm}
\usepackage{amssymb}
\usepackage{amsmath}
\usepackage{multirow}

\newcommand{\ignore}[1]{}
\newcommand{\vect}[1]{\mathbf{#1}}

\usepackage{ntheorem}
\theoremseparator{:}
\newtheorem{hyp}{Hypothesis}

\newcommand{\subparagraph}{}
\renewcommand{\baselinestretch}{0.96}
\title{\LARGE \bf AG2U - Autonomous Grading Under Uncertainties }

\author{Yakov Miron$^{1,2}$, Yuval Goldfracht$^{1}$, Chana Ross$^{1}$, Dotan Di Castro$^1$ and Itzik Klein$^2$\\ 
\thanks{$^1$Bosch Center for Artificial Intelligence (BCAI), Haifa, Israel}
\thanks{$^2$The Autonomous Navigation and Sensor Fusion Lab, The Hatter Department of Marine Technologies, University of Haifa, Israel} 
{\tt\small {\{Yakov.Miron, Yuval.Goldfracht, Chana.Ross, Dotan.DiCastro\}}@bosch.com}, 
{\tt\small kitzik@univ.haifa.ac.il }
}

\makeatletter
\def\endthebibliography{%
	\def\@noitemerr{\@latex@warning{Empty `thebibliography' environment}}%
	\endlist
}
\let\NAT@parse\undefined
\makeatother

\usepackage[colorlinks,citecolor=red,urlcolor=blue,bookmarks=false,hypertexnames=true]{hyperref}

\renewcommand\footnotemark{}

\begin{document}
	
	\maketitle

	\thispagestyle{plain}
    \pagestyle{plain}

    \begin{abstract}

Surface grading, the process of leveling an uneven area containing pre-dumped sand piles, is an important task in the construction site pipeline.
This labour-intensive process is often carried out by a dozer, a key machinery tool at any construction site.
Current attempts to automate surface grading assume perfect localization. However, in real-world scenarios, this assumption fails, as agents are presented with imperfect perception, which leads to degraded performance. 
In this work, we address the problem of autonomous grading under uncertainties. First, we implement a simulation and a scaled real-world prototype environment to enable rapid policy exploration and evaluation in this setting. Second, we formalize the problem as a partially observable markov decision process and train an agent capable of handling such uncertainties. We show, through rigorous experiments, that an agent trained under perfect localization will suffer degraded performance when presented with localization uncertainties. However, an agent trained using our method will develop a more robust policy for addressing such errors and, consequently, exhibit a better grading performance.

\end{abstract}

    \section{Introduction} \label{Introduction}

Recent years have seen a growing demand for automation at construction sites. First, automation can reduce the amount of manual labor required from construction workers, thus helping resolve this industry's noted labor shortage. Second, it can increase productivity, which has been fairly stagnant in the last few decades, and cut down the inflating costs. Lastly, it can improve worker safety by allowing machines to handle risky tasks.  

Yet, incorporating automation into construction sites is a complicated endeavor due to the unpredictable, unstructured nature of these environments, where multiple machines and workers work simultaneously on a variety of challenging tasks. Construction projects are also extremely varied, with each one tailored to a specific architectural design, specifications etc. For these reasons, achieving automation in construction sites is an extremely difficult task, which has not yet been solved \cite{pradhananga2021identifying}. 

Recent advancements in the field of artificial intelligence (AI), in the context of autonomous vehicles, hold promise for automation in the construction industry. While indeed related, the automotive and construction industries nevertheless pose different challenges for automation. For example, data collection, which is the backbone of current methods for autonomous driving, is extremely challenging in the unstructured environment of construction sites, where safety, time, and cost are major practical considerations. This problem can be partially solved using simulators, but these, too, have their own set of drawbacks. In addition, the unpredictability of the construction environment, where extreme and dangerous scenarios happen frequently, has been found to be difficult to model and learn using standard methods for autonomous vehicles.

In this work, we address the problem of autonomous path planning for construction site vehicles. Specifically, our focus is on the autonomous grading task done by the dozer under localization uncertainty, where the estimated pose of the vehicle is erroneous. This task poses several challenges common to all machinery tools in any construction site. Therefore, the addressed problem can be considered a representative example in the field. The main challenges are data collection, which is a key difficulty for all machinery tools, partial observability of the environment due to sensor positioning, and sensory noise that causes localization uncertainty. The latter affects the way the agent perceives the environment and, thus, significantly impedes the decision-making process. 

In order to overcome the difficulty of data collection, we use domain adaptation techniques \cite{jiang2021simgan, zhao2020sim} that can bridge the sim-to-real gap. In our approach, we augment the simulation so as to resemble as much as possible real-world data. We then train (and evaluate) a learned policy purely in simulation and test it in both simulation and in a scaled prototype environment. To overcome the localization uncertainty, we devise a novel training regime where the uncertainty is taken into account during the agent's policy training. This allows the agent to learn a robust policy, which improves its performance under uncertainty during inference compared to an agent  trained in a clean, noise-less environment.
Specifically, we augment the training dataset with many variations, including scale, rotation, and translation versions of the observation, thus improving the agent's ability to cope with more realistic scenarios where the observation is uncertain due to localization errors. 

\noindent Our main contributions are as follows: 
\begin{enumerate}
    \item We show that an agent trained to perform a grading task in a perfect localization setting
    will under-perform when presented with 
    real-world uncertainties.
    
    \item We propose a training regime that takes the sensory noise into account to produce a robust policy in the presence of uncertainties.
    
\end{enumerate}
\noindent We prove our hypotheses on both a simulation setup and a unique scaled real-world environment setup that includes a dozer with relevant sensors.

    \section{Related Work} \label{related_work}

\subsection{Autonomous Vehicle Localization} \label{subsec:autonomous_vehicles_localization}

Localization, i.e., estimating the current position and attitude of the vehicle with respect to its surroundings, is key to achieving safe, reliable and robust decision-making in autonomous driving (AD). While localization has been extensively studied in the broad field of AD, in the context of autonomous construction vehicles it has received far less attention. AD can be broadly split into two types of scenarios: driving in urban areas and on highways. There are three common approaches for tackling urban areas: \textit{(a)} 3D registration-based methods, which fuse offline 3D maps to current Lidar scans \cite{tam2012registration}. \textit{(b)} 3D feature-based methods, such as \cite{mur2017orb}, which design relevant features in the 3D space that are then used to calculate the displacements between successive scans. \textit{(c)} 3D deep learning-based methods, where 2D camera images are used to predict the odometry between a pair of images for later processing \cite{clark2017vinet, wang2017deepvo, yang2018deep}. When it comes to highway settings, where the high velocity of the vehicle impinges upon the performance of the Lidar, \cite{svensson2016ego}, \cite{hata2014road} suggest to use road markings and information from surrounding vehicles.


Localization methods often rely on the fusion of multiple noisy sensor measurements in order to produce a more accurate estimation of the vehicle's state. A common approach is the extended Kalman filter (EKF), a nonlinear version of the well-known linear Kalman filter, which linearizes about an estimate of the current mean and covariance. In the case of well-defined transition models, the EKF is considered the standard in the theory of nonlinear state estimation, navigation systems and GPS
\cite{noureldin2013fundamentals}.
A possible extension of the standard EKF is error-state EKF (ES-EKF) \cite{roumeliotis1999circumventing, madyastha2011extended}, where the estimated state is the error between the current prediction and the measurement. This variant of the EKF assumes that the errors have a less complex behaviour than the state itself and, therefore, the linearization is more accurate. In this setting, an aiding (primary) sensor that produces fairly accurate but low-frequency measurements, given in the \textit{navigation} coordinate system, is used alongside a high-frequency but less accurate sensor in order to produce a high-frequency and accurate estimation of the state \cite{noureldin2013fundamentals}.

\subsection{Autonomous Vehicles' Decision-Making} \label{subsec:autonomous_vehicles_decision_making}

Research in the field of decision-making for autonomous construction vehicles is somewhat limited and mainly focused on the analysis of vehicle dynamics. \cite{taghavifar2017off} present an extensive analysis of the forces and moments applied to vehicles in off-road settings. In contrast, research on autonomous vehicles in urban surroundings is abundant. \cite{gao2020vectornet} use a graph neural network (GNN) that exploits the spatial locality of individual road components, thus offering better scene understanding for decision-making. Moreover, \cite{romero2021maneuver} optimize the driving comfort and travel duration while staying within the safety limits. To do so, they combine continuous planning with semantic information. This allows the planning system to deal with the complexity of the problem in a computationally efficient manner.
\cite{bansal2018chauffeurnet} propose to apply data augmentation, i.e., add perturbations to the labeled trajectory, as part of the training process. This method allows one to augment interesting scenarios, such as collisions and/or lane divergence, that are missing from the original labeled data. This process leads to a more robust policy that can handle diverse scenarios and overcome the distribution shift between the simulation and the real world. This approach proved to be very efficient for agents' robustness to perception noise and was, therefore, adopted in this work.


\subsection{Sand Simulation} \label{subsec:related_work_sand_simulation}
In the field of autonomous vehicles, great effort is devoted to the correct modelling of the interaction between the vehicle and its environment. Understanding and modelling this interaction is crucial for learning a good policy that will succeed in real-world settings. In the context of construction sites, in general, and dozers, specifically, a key factor in the behavior of the vehicle is its interaction with the soil, i.e., the sand is moved by the vehicle as it travels. Precise particle simulation can be used to model a vehicle's interaction with the sand; this approach involves using solid mechanical equations \cite{solid_mechanics}, discrete element methods \cite{finite_elements}, or fluid mechanics \cite{SULSKY1994179}. Recently, \cite{sanchez2020learning} utilized deep neural networks to simulate the reaction of particles to forces. While such methods might provide accurate modelling and realistic visualization, they often entail high computational costs. In the context of learning a policy, rapid data collection and evaluation are essential---even at the expanse of simulation accuracy. We, therefore, must be able to capture the essence of the interaction, as doing so will allow the agent to succeed in the real-world without compromising the rendering speed.  \cite{ross2021agpnet} established a simulator for earth-moving vehicles, where the goal was to enable policy evaluation for bulldozers. This simulation was fairly accurate, fast and able to capture the main aspects of the interaction between the sand and the vehicle. 

\subsection{Autonomous Construction Vehicles} \label{subsec:autonomous_construction_vehicles}
The area of autonomous construction vehicles has enjoyed a growing interest over the past few years. In this field, the majority of work has focused on bulldozers, excavators and wheel-loaders in multiple aspects of the autonomous driving task. \cite{hirayama2019path} implemented a heuristic approach to grade sand piles. They examined the trade-off between grading the pile when the blade is in full capacity and pushing less sand to reduce the elapsed time. \cite{ross2021agpnet} presented a ML approach for autonomous grading that exhibited better generalization capabilities in previously unseen scenarios, both in simulation and in real-world settings. Moreover, \cite{miron2022towards} trained a privileged agent to mitigate the sim-to-real perception gap using a simulator.
\cite{stentz1999robotic}, developed a pioneering software package on a construction vehicle for automation; their excavator was one of the first to operate autonomously. \cite{kurinov2020automated} trained an agent using a model-free Reinforcement Learning approach. They relied on the PPO-CMA algorithm and a semi-recursive multi-body method and showed impressive results in simulation. Moreover, \cite{oh2015integrated} offer a simulator for improving the performance and energy flow of wheel loaders. However, none of the presented methods takes into account the sensory noise of the system. This fact hinders the efforts to deploy these methods in a real-world environment.  
\subsection{POMDP} \label{subsec: POMDP}
A partially-observable markov decision process (POMDP; \cite{sutton2018reinforcement}) consists of the tuple $(\mathcal{S}, \mathcal{O}, \mathcal{A}, \mathcal{P}, \mathcal{R})$. The state $s \in \mathcal{S}$ contains all the information required to learn an optimal policy. However, agents are often provided with partial or noisy information regarding the environment, which is termed observation $o \in \mathcal{O}$. 
As opposed to states, observations typically lack the sufficient statistics for optimality. At each state $s$, the agent takes an action $a \in \mathcal{A}$. Then, the system transitions to the next state $s'$ based on the transition kernel $P(s' | s, a)$. Finally, the agent is provided with a reward $r(s, a)$. The goal of an agent is to learn a policy $\pi$ that maximizes the cumulative \textit{reward-to-go}, where the policy maps from the observations (or estimated states) to the actions.

\subsection{Dataset Augmentation}\label{subsec:Dataset Augmentation}
Dataset augmentation is a common method and best practice for training ML models \cite{Ross2011ARO, shorten2019survey}. If the model takes images as input, common augmentations often include basic image manipulations such as scale, rotation, translation, random erasing, color space changes (brightness changes etc.). In the context of decision-making, augmentations can help reduce the distribution mismatch between simulation and real-world samples \cite{bansal2018chauffeurnet}. Domain adaptation \cite{jiang2021simgan, miron2019s} aims to improve the simulation's photo-realism in order to minimize the sim-to-real perception gap. 



    \section{Autonomous Grading Under Uncertainties} \label{Methods}

In this work, we examine the effect of uncertainties in pose estimation on the performance of agents with respect to the grading task. In our approach, we model this uncertainty as noisy observations of the true state, as described in Section \ref{subsec: Simulating_Uncertainties}. Inspired by \cite{bansal2018chauffeurnet}, we hypothesize the following:
\begin{hyp} \label{Hypothesis_1}
An agent trained under a perfect localization setting $(agent_1)$ will underperform when presented with real-world uncertainties at inference.
\end{hyp}
\begin{hyp} \label{Hypothesis_2}
An agent trained with noisy observations $(agent_2)$ will develop a robust policy that can handle noisy observations at run-time.
\end{hyp}

\begin{figure*}[ht!]
    \centering
     \includegraphics[width=0.95\textwidth]{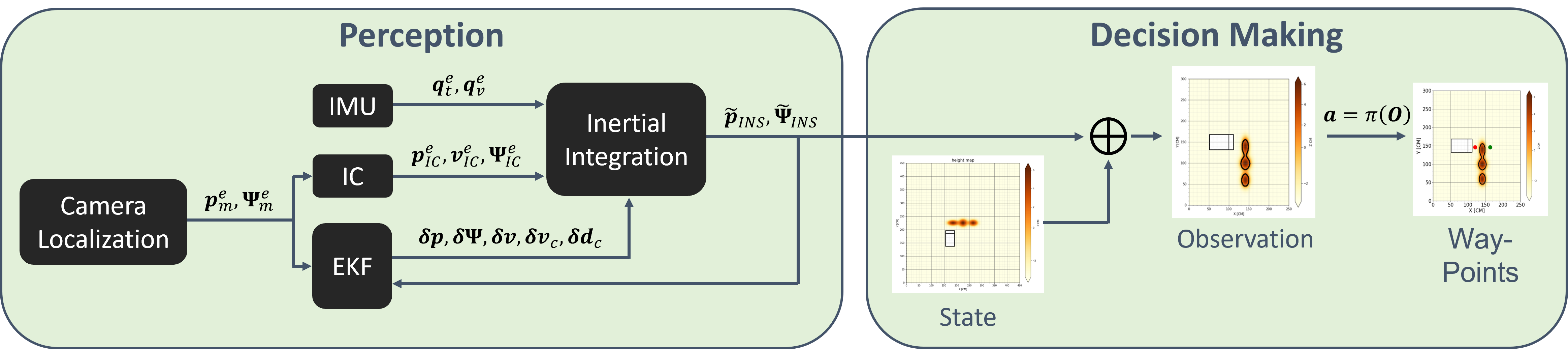}\hfill
     \caption{
     Overview of the suggested method: 
     \textbf{Perception}:
        An IMU provides velocity and angular rate increments  ($\vect{q}^e_{v}, \vect{q}^e_{t}$, respectively) at high frequency. The camera provides the aiding position and attitude measurements ($\vect{p}_{m}^e, \vect{\Psi}_{m}^e$, respectively) at low frequency. The inertial integration algorithm uses both IC and IMU measurements to produce position and attitude estimates at high frequency  ($\Tilde{\vect{p}}_{INS}$, $\Tilde{\vect{\Psi}}_{INS}$, respectively). Once an aiding measurement is available (at 1 Hz in simulation), the EKF is activated and provides corrected pose and bias estimates. These estimates are fed back to the inertial integration algorithm for increment compensation. The output of the perception block is an  estimate of the pose at high frequency.
        \textbf{Actuation}: The estimated pose from the perception block is fed to the simulator in order to render an observation from the true state and the estimated pose. Once the observation is available, it is fed to the policy $\pi$, which provides way-point decisions. The simulator then performs these actions, and this process is repeated.
    }

     \label{fig: overall_system}
\end{figure*}

\noindent In this section, we discuss in detail the proposed method for training an agent under uncertainties for the purpose of learning a robust policy in real-world settings. Figure \ref{fig: overall_system} provides an overview of our perception and decision-making pipeline, as described below.

\subsection{Problem Formulation} \label{subsec: grading as mdp}
In order to tackle the task of autonomous grading, we formalize it as a POMDP/R and define a 4-tuple consisting of states, observations, actions, and the transition kernel, as discussed in Section \ref{subsec: POMDP}.

\begin{figure}
    \centering
    \subfloat[]{\includegraphics[width=0.37\linewidth]{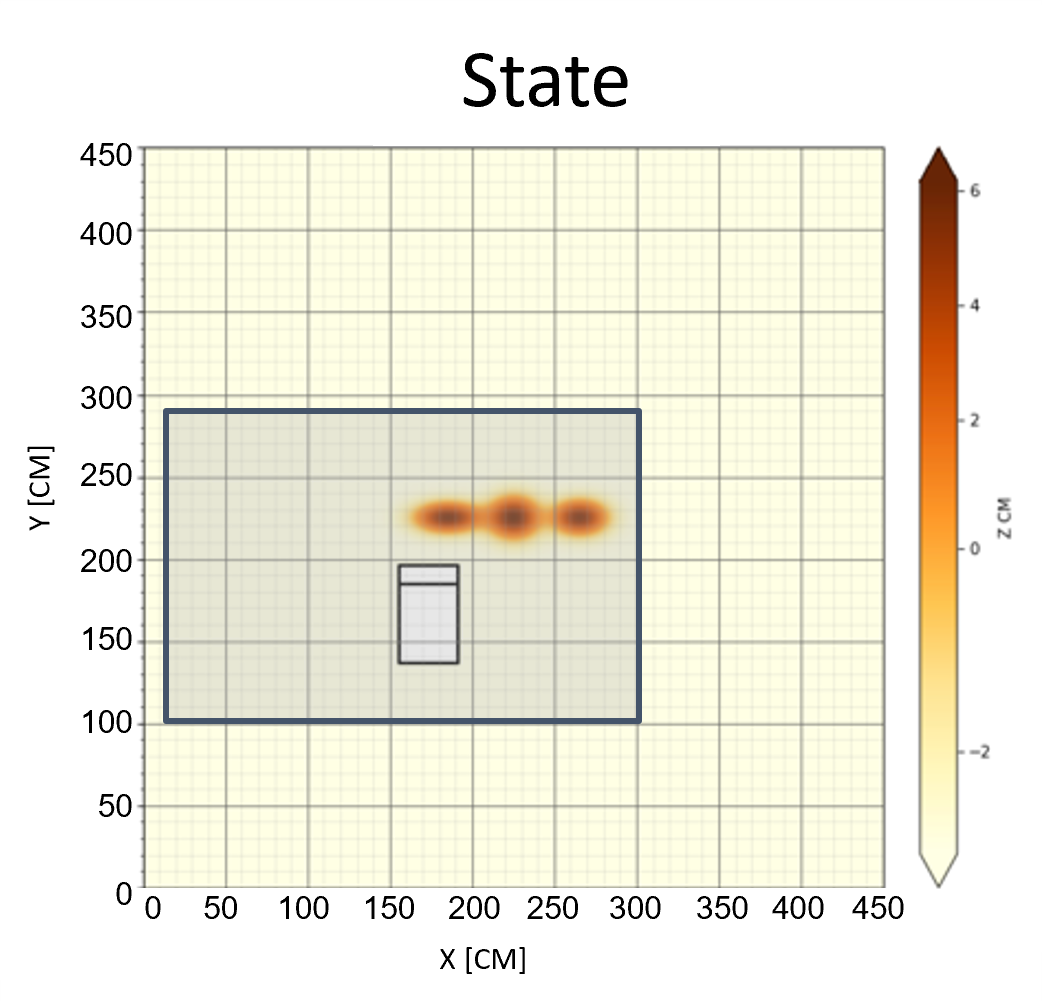} \label{fig:state_obs_action_s}} 
    \subfloat[]{\includegraphics[width=0.32\linewidth]{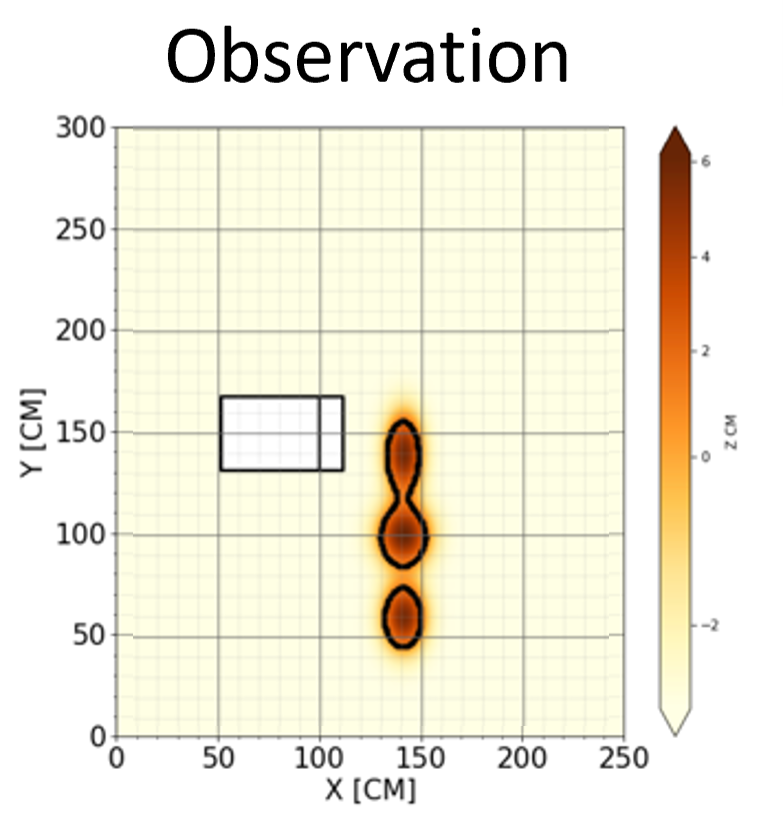} \label{fig:state_obs_action_o}}
    \subfloat[]{\includegraphics[width=0.32\linewidth]{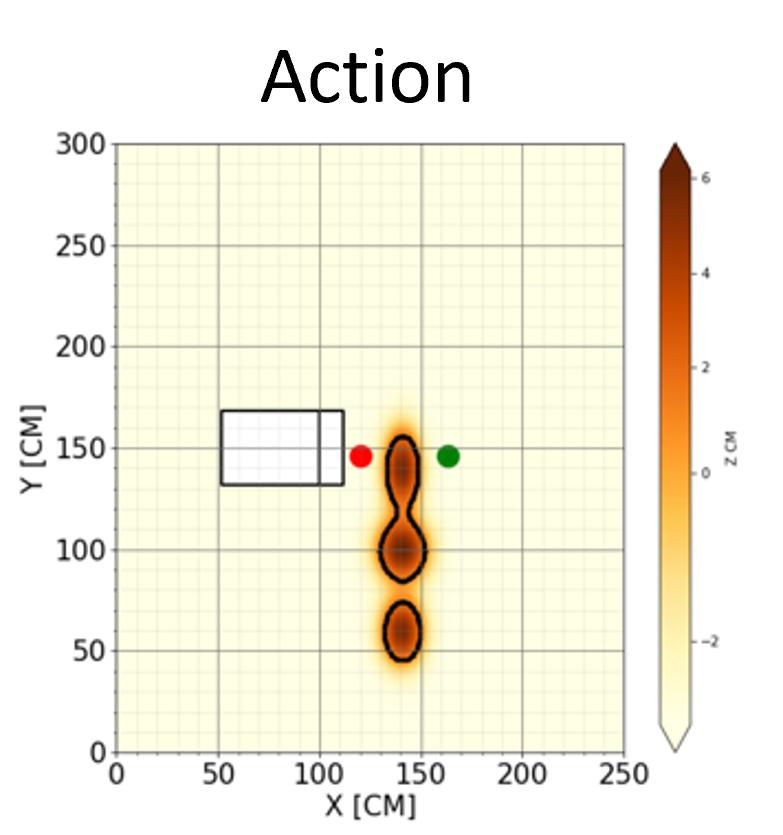} \label{fig:state_obs_action_a}}
    
    \caption{Visualization of state, observation and actions. \textbf{(a)} The \textit{full state}, where the agent has access to all the information, without errors. \textbf{(b)} \textit{Observation}, where the agent has access to \textit{part} of the information, which may include errors because the translation from state to observation (see section \ref{subsec: POMDP}) uses the estimated pose. The observation is derived from the grey rectangle of the state \textbf{(c)} \textit{Actions}, i.e., the decisions made by the agent, given an observation, described as way-points, to reach and reverse (green and red dots respectively).}
    \label{fig:state_obs_action} 
\end{figure}

\noindent \textbf{States:} The state consists of all the information required to obtain the optimal policy and determine the outcome of each action. In our case, the state includes the accurate pose of the agent (see Figure \ref{fig:state_obs_action_s}).
\noindent \textbf{Observations:} 
An observation aims to reflect the state's uncertainty caused by the sensor's noise. In this case, the inaccurate state (mainly pose) estimation translates into an erroneous bounding box view around the current location of the dozer. In the simulation, we mimic this behavior by applying augmentation (rotation, translation) to the true, accurate state. Figure \ref{fig:state_obs_action_o} is a case where an observation is derived from the true state without errors. 
\noindent \textbf{Actions:} We consider two aspects of the action in the context of errors: (i) \textit{open-loop}, where the policy outputs a way-point for the agent to reach. Here, pose estimation errors are presented as a sub-optimal projection from state to observation. (ii) \textit{closed-loop}, where errors in pose-estimation are fed back to the low-level controller for trajectory execution. Here, errors propagate through the system, leading to divergence from the desired path. Figure \ref{fig:state_obs_action_a} includes examples of selected actions, in the form of way-points, shown as green and red dots.
\noindent \textbf{Transitions:} Transitions are determined by a dozer's dynamics and the physical properties of the soil and the environment.

\subsection{Training Under Uncertainties}
\begin{figure}[ht]
    \centering
        \includegraphics[width=.99\linewidth]{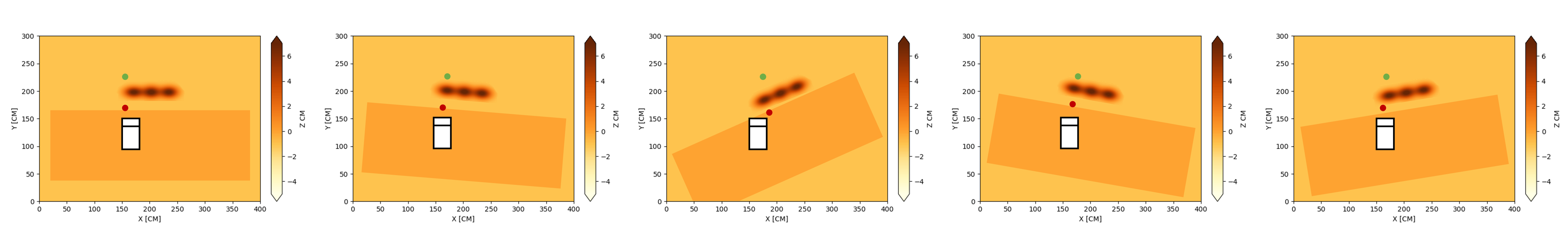}    
        \caption{
        Visualization of the augmented training dataset. The left image presents the true pose, i.e., the true observation. The green and red dots are the actions' training labels. The remainder images are rendered observations from the distribution of the EKF estimation, and 
        their predicted training labels, in the form of way-points, shown as green and red dots.}
    \label{augmented_dataset}
\end{figure}

\noindent In order to validate our hypothesises, we train \textit{two} agents, each one under one of \textit{two} noise settings:
 \label{three_types_of_experiments}
\noindent \textbf{Noise-free}: In this setting, we use the true trajectory, within the simulation, in order to extract noise-free observations of the current state. This scenario serves as a baseline for future comparisons. We consider the actions taken by the agent under this setting as the optimal policy. 
\noindent \textbf{Noise with Sensor Fusion Filtering}: 
In this setting, we render many noisy observations (augmentations) generated by the sensor fusion filtering algorithm. We do so by (i) adding synthetic noise to the inertial sensors, arranged in a typical inertial measurement unit (IMU), and to the aiding sensor measurements, (ii) applying our inertial navigation system (INS) and EKF equations as described in section \ref{subsec: Simulating_Uncertainties} and Figure \ref{fig: overall_system}, (iii) rendering the noisy observations from the distribution produced by the filter. These actions introduce uncertainties into our training pipeline.
\noindent We denote the agents trained using the observations derived from the (i) noise-free and (ii) noise-with-sensor-fusion-filtering settings as $agent_1$, $agent_2$, respectively. In practice, inserting sensory noise into our measurements translates into small perturbations around the original observations. This process enhances our training dataset, which now includes a much wider distribution of potential states. This, in turn, allows the agent to learn a policy that is more robust to localization uncertainty.
Figure \ref{augmented_dataset} provides a visualization of the augmented dataset for training, where possible observations are rendered from the estimated pose $\Tilde{\vect{x}}$, while taking into account the pose uncertainty (from the EKF covariance matrix estimate $\vect{\Sigma}$), i.e.:
\begin{equation}  \label{eq: observation_rendering}
    \{\Tilde{\vect{x}}_k\}_{k=0}^{K-1} \sim \mathcal{N}(\Tilde{\vect{x}} ,\,\vect{\Sigma})\ 
\end{equation}
Here, $K$ is the number of observations that were rendered from this distribution about the estimated pose $\Tilde{\vect{x}}$, and $\mathcal{N}(\cdot, \cdot)$ is the normal distribution, 

\subsection{Navigation Filter}\label{subsec:SD algorithm Our}

An INS uses its inertial sensor readings and initial conditions (IC) to calculate the platform's position, velocity, and orientation~\cite{noureldin2013fundamentals}. In our implementation, we employ a sensor fusion approach, using EKF, between the IMU readings and an additional aiding sensor. When applying inertial integration to the IMU readings, we can neglect the effect of Earth's rotation, as the episode time is short \cite{noureldin2013fundamentals}. Within the INS algorithm, we approximate the derivative as a temporal difference and, therefore, use Euler integration \cite{ravat1996analysis}, rather than trapezoidal integration, which is commonly used in numerical integration \cite{ravat1996analysis} and inertial navigation \cite{noureldin2013fundamentals}. In practice, we make use of the angular \eqref{eq: ang_inc_equ} and velocity \eqref{eq: vel_inc_equ} increments at time $k$ from the inertial sensor.

\begin{subequations} \vspace{-0.3cm}
\small
\begin{align}
    \vect{q}_t[k]\in\mathbb{R}^{3 \times 1} \label{eq: ang_inc_equ}\\
    \vect{q}_v[k]\in\mathbb{R}^{3 \times 1} \label{eq: vel_inc_equ}
\end{align} \label{eq: increments equation}
\normalsize
\end{subequations} \vspace{-0.3cm}

\noindent In addition, we define the $E2D$ operator as follows:
\begin{equation}  \label{eq: E2D}
    \small
    \vect{D} := \vect{F}(\vect{q}_t[k]) = E2D(\vect{q}_t[k])\in\mathbb{R}^{3 \times 3}
    \normalsize
\end{equation}
This operator describes the deterministic transformation  from Euler angels ($\psi, \theta, \phi$) to a transformation matrix $\vect{D}$, referred to hereunder as $\vect{F}(\cdot)$.
The following equations describe our implementation of the inertial integration: 

\begin{subequations} \vspace{-0.3cm}
\small
\begin{align}
    \vect{D}_n^b[k] &= \vect{F}(\vect{q}_t[k]) \cdot{\vect{D}_n^b[k-1]} \label{eq: att_equ} \\
    \Tilde{\vect{v}}^n[k] &= \Tilde{\vect{v}}^n[k-1] + \vect{D}_b^n[k]\cdot{\vect{q}_v[k]} + (0,0,g\cdot{dt})^T \label{eq: vel_equ} \\ 
    \Tilde{\vect{p}}^n_{INS}[k] &= \Tilde{\vect{p}}_{INS}^n[k-1] + \Tilde{\vect{v}}^n[k]\cdot{dt} \label{eq: pos_equ}
\end{align} \label{eq: sd equation}
\normalsize
\end{subequations}

\noindent In \eqref{eq: att_equ}, $\vect{D}_n^b[k]$ denotes the attitude transformation matrix between the navigation and body frames at time $k$, i.e., after integration and $\vect{D}_n^b[k-1]$ before integration. 
In \eqref{eq: vel_equ}, $\Tilde{\vect{v}}^n[k], \Tilde{\vect{v}}^n[k-1]$ are the current and previous estimated velocities, respectively, $\vect{D}_b^n[k]$ is the transformation matrix from body to navigation frame, where $\vect{D}_b^n[k] = (\vect{D}_n^{b}[k])^T$, $g$ is Earth's gravitational constant, and $dt$ is the sampling frequency of the inertial sensor.
In \eqref{eq: pos_equ}, $\Tilde{\vect{p}}_{INS}^n[k],     
\Tilde{\vect{p}}_{INS}^n[k-1]$ are the current and previous estimated positions, respectively.

\noindent Our sensor fusion algorithm relies on the error-state extended Kalman filter (ES-EKF) to fuse pose estimates from the INS integration algorithm with the low-frequency pose measurements from the aiding sensor (see section \ref{subsec:SD algorithm Our}). In simulation, our aiding sensor measurements are derived from the generated trajectory pose, which can be either noiseless or contain synthetically added noise. In our scaled real-world environment, the aiding sensor measurements are estimated using an ArUco marker \cite{romerospeeded} captured by our perception setup (see Figure \ref{fig: Lab_experimental_setup}). These measurements are inherently noisy and can, therefore, approximate a real-world experiment. In addition, we solely model the constant bias of the IMU, neglecting other terms (e.g., in-run stability, since for our short time episodes, these terms are negligible). In our implementation, we define the error-state and measurement vectors as:

\begin{subequations} \vspace{-0.3cm} \label{eq: error state equations}
\small
\begin{align}
    \vect{\delta x} &= [\vect{\delta p},  \vect{\delta \Psi}, \vect{\delta v}, \vect{\delta b}_c, \vect{\delta d}_c]^T   \in\mathbb{R}^{15\times 1} 
    \label{eq: error-state vector imp}  \\
    \vect{\delta z} &= [\vect{\delta p}_{m}, \vect{\delta \Psi}_{m}]^T \in\mathbb{R}^{6\times 1}  
    \label{eq: error-state measurement imp} 
\end{align}
\normalsize
\end{subequations}
In \eqref{eq: error-state vector imp}, $\vect{\delta x}$ is the error-state vector, $\vect{\delta p}$ is the position error, $\vect{\delta \Psi}$ is the attitude error, $\vect{\delta v}$ is the velocity error, and $\vect{\delta b}_c, \vect{\delta d}_c $ are the constant accelerometer and gyroscope bias, respectively. In \eqref{eq: error-state measurement imp}, $\vect{\delta z}$ is the difference between the measurement from the aiding sensor, provided by the perception setup, and the estimate from the INS algorithm, i.e., $\vect{\delta z}$ is the measurement residual. In our setup, the aiding measurement includes the position and attitude, where $\vect{\delta p}_{m} = \vect{p}_{m}^e - \Tilde{\vect{p}}_{INS}^n$ is the positional measurement difference, $\vect{p}_{m}^e$ is the noisy position measurements from the aiding sensor, and $\Tilde{\vect{p}}_{INS}^n$ is the noisy position estimate, according to \eqref{eq: pos_equ}.
$\vect{\delta \Psi}_{m}$ is the attitude measurement residual, according to \cite{noureldin2013fundamentals}:
\begin{equation}  \label{eq: delta_psi_equation}
    \small
        \vect{\delta \Psi_{m}} = \vect{h}(\vect{F}(\vect{\Psi}^e _{m})^T \cdot \vect{F}(\Tilde{\vect{\Psi}} _{INS})) \in\mathbb{R}^{3 \times 1}
    \normalsize
\end{equation}
Here, $\vect{\Psi}^e _{m}$ and $\Tilde{\vect{\Psi}} _{INS}$ are the noisy attitude measurements from the aiding sensor and INS algorithm, respectively, $\vect{F}(\cdot)$ is the $E2D$ operator defined in \eqref{eq: E2D}, and $\vect{h}(\cdot)$ is the $D2E$ operator, which represents a deterministic function from a transformation matrix $\vect{D}$ to Euler angels ($\psi, \theta, \phi$), defined as follows:
\begin{equation}  \label{eq: D2E}
    \small
    (\psi, \theta, \phi) := \vect{h}(\vect{D}) = D2E(\vect{D}) \in\mathbb{R}^{3 \times 1}
    \normalsize
\end{equation}
$\Tilde{\vect{\Psi}} _{INS}$ is derived when applying the $\vect{h}(\cdot)$ operator to $\vect{D}_n^b[k]$ in \eqref{eq: att_equ}.
The intuition behind \eqref{eq: delta_psi_equation} is that the product of $\vect{F}^T \cdot \vect{F}$ matrices corresponds to a difference operator in linear (Euler angles) space. The system matrix of our ES-EKF implementation is as follows:

        

\begin{equation}
\small
  \vect{A} = 
        \begin{bmatrix}
        \vect{I}_3                & \vect{0}_3 & \vect{I}_3\cdot dt_m & \vect{0}_3       & \vect{0}_3       \\
        \vect{0}_3                & \vect{I}_3 & \vect{0}_3    & \vect{0}_3       & -\Sigma \vect{D} \\
        \vect{I}_3 \cdot \frac{1}{dt_m} & \vect{A}_s & \vect{I}_3    & \Sigma \vect{D}  & \vect{0}_3\\
        \vect{0}_3                & \vect{0}_3 & \vect{0}_3    & \vect{I}_3       & \vect{0}_3\\
        \vect{0}_3                & \vect{0}_3 & \vect{0}_3    & \vect{0}_3       & \vect{I}_3\\
        \end{bmatrix}  \in\mathbb{R}^{15\times 15}
\normalsize        
\label{eq: error-state ekf model imp} \\
\end{equation} 
Here, $\vect{I}_3, \vect{0}_3$ are a $3 \times 3$ identity and zeros matrix, respectively, $\Sigma \vect{D}$ is the sum of the rotation matrix between the measurement periods $k$ and $k+1$, i.e., $ \Sigma \vect{D} = \sum_{l=k}^{l=k+1} \vect{D}[l]  $, $\vect{A}_s$ is a skew symmetric matrix representing the difference between two velocity measurements, and $dt_m$ is the time interval between measurements $k$ and $k+1$. In addition, as we update the position and attitude within the EKF, we define the following measurement matrix:
\begin{equation}
\small
    \vect{H} = 
        \begin{bmatrix}
        \vect{I}_3 & \vect{0}_3 & \vect{0}_3 & \vect{0}_3 & \vect{0}_3       \\
        \vect{0}_3 & \vect{I}_3 & \vect{0}_3 & \vect{0}_3 & \vect{0}_3       \\
        \end{bmatrix} \in\mathbb{R}^{15\times 6}
        \label{eq: error-state ekf proj imp}           
\normalsize        
\end{equation} 

    \section{Analysis and Results}

\subsection{Simulating Uncertainties} \label{subsec: Simulating_Uncertainties}
We used the simulation described in \cite{ross2021agpnet} as a starting point, and made considerable modifications to it in order to support localization errors due to sensory noise. The simulation provides a full trajectory, i.e., positions and attitudes, at the beginning and end of each leg. In order to simulate the localization uncertainties in a realistic manner, we added the ability to generate a full trajectory at high frequency, e.g., 100 Hz. These high-frequency measurements allow us to simulate the sampling rate of the inertial sensor. From these high-rate measurements, we sample the low rate, e.g., 1 to 10 Hz, aiding measurements required for the sensor fusion algorithm. 

In practice, we generate a translation trajectory using a linear interpolation of the shortest straight path between the initial and terminal locations in a grading leg. Similarly,  for a rotation trajectory, we interpolate between the initial and terminal attitudes of each leg (see Algorithm \ref{Algorithm:Attitude Interpolation Algorithm}). We consider this trajectory a true trajectory executed by a dozer. We then apply a second-order derivative to the true positions and a first-order derivative to the attitude measurements in order to retrieve the velocity and rotation increments, respectively \cite{noureldin2013fundamentals}. These measurements are considered the clean inertial readings (see Algorithm ~\ref{Algorithm:IMU increments generation}), and are given in body coordinates. Building off the clean inertial increments, we can then add sensor noise in order to simulate real measured inertial increments such as:

\begin{subequations}  \label{eq: IMU noise} 
\small
\begin{align}
\vect{q}_v^e &= \vect{q}_v^t + \vect{b}_c \Delta t + \vect{a}_{rw} \sqrt{\Delta t} \in\mathbb{R}^{3 \times 1} \\
\vect{q}_t^e &= \vect{q}_t^t + \vect{d}_c \Delta t + \vect{g}_{rw} \sqrt{\Delta t} \in\mathbb{R}^{3 \times 1} 
\end{align}
\normalsize
\end{subequations}
where $\vect{q}_v^e$ is the erroneous version of $\vect{q}_v^t$, $\vect{b}_c$ is the constant bias of the accelerometer, and $\vect{a}_{rw}$ is the accelerometer random walk. Similarly, $\vect{q}_t^e$ is the erroneous version of $\vect{q}_t^t$, $\vect{d}_c$ is the constant bias of the gyroscope, and $\vect{g}_{rw}$ is the gyroscope random walk. In addition, IC errors are applied to the true position and velocity vectors \small $\vect{p}_{IC}^t, \vect{v}_{IC}^t$ \normalsize, respectively:

\begin{subequations}  \label{eq: IC noise} 
\small
\begin{align}
    \vect{p}_{IC}^e &= \vect{p}_{IC}^t + \vect{\Delta p}_{IC} \in\mathbb{R}^{3 \times 1} \\ 
    \vect{v}_{IC}^e &= \vect{v}_{IC}^t + \vect{\Delta v}_{IC} \in\mathbb{R}^{3 \times 1} 
\end{align}
\normalsize
\end{subequations}
where \small $\vect{\Delta p}_{IC}$ \normalsize and \small $\vect{\Delta v}_{IC}$ \normalsize are the initial position and velocity errors, respectively. In the case of attitude, as \small $\vect{\Psi}_{IC}^t$ \normalsize is the true attitude, the following formulation was used:
\begin{equation} \label{eq: att noise}
\small
\vect{\Psi}_{IC}^e = \vect{h}( \vect{F}(\vect{\Psi}_{IC}^t) \cdot{\vect{F}(\vect{\Delta \Psi}_{IC})} ) \in\mathbb{R}^{3 \times 1} 
\normalsize
\end{equation}

\noindent where \small $\vect{\Delta \Psi} _{IC}$ \normalsize is the initial attitude error.
Moreover, noise is added to the true position and attitude aiding measurements according to the following equations:

\begin{subequations}  \label{eq: primary sensor noise in sim} 
\small{
\begin{align}
\vect{p}_{m}^e &= \vect{p}_{m}^t + \vect{\Delta p}_{err} \in\mathbb{R}^{3 \times 1}  \\ 
\vect{\Psi}_{m}^e &= \vect{h}(\vect{F}(\vect{\Psi} _{m}^t)^T \cdot{ \vect{F}(\vect{\Delta \Psi} _{err})}) \in\mathbb{R}^{3 \times 1} 
\end{align}
}
\normalsize
\end{subequations}

\noindent where $\vect{p}_{m}^e, \vect{\Psi}_{m}^e$ are the erroneous, synthesized position and attitude measurements of the aiding sensor, respectively, $\vect{p}_{m}^t, \vect{\Psi} _{m}^t$ are the true position and attitude measurements, respectively, and $\vect{\Delta p}_{err}, \vect{\Delta \Psi} _{err}$ are the position and attitude errors added to the simulation, respectively.

As our policy model, we used a ResNet-based \cite{he2016deep}, end-to-end, fully convolutional network with 8 layers, all with 64 channels, dilated convolutions with a dilation factor of 2 and Relu activations. The input size is \small$H\times W$, \normalsize and the output size is \small$H\times W\times2$ \normalsize---for the desired way-point and the next iteration way-point, respectively (green and red marks in Figure \ref{fig:decisions_real_exp}).  In addition, we train using an ADAM optimizer \cite{kingma2014adam} with a learning rate of $0.001$. $agent_2$ was trained using 200 random initial states (episodes) for 500 epochs. In each episode, we randomize the number of sand piles, shapes, locations, and volumes and the initial location of the dozer.

\newcommand{\forcond}{$k=0$ \KwTo $(n_{int}-1)$}
\begin{algorithm}[ht]
\DontPrintSemicolon
\KwInput{\small{$\vect{q}_n^1, \vect{q}_n^2$} are two independent quaternions}
\KwInput{$n_{int}$ is the number of interpolation points}
\KwOutput{interpolated quaternions 
\small{$\{\vect{q}_{interp}[k]\}_{k=0}^{n_{int}-1}$}
}

$\otimes$ represents quaternion multiplication \\
$ \vect{q}=[r,\vect{i}]^T \in\mathbb{R}^{4 \times 1}  $  represents the stacking of the real and imaginary parts to a quaternion.\\
    $\vect{q}_{inc}$ = $\frac{\vect{q}_n^1 \otimes \vect{q}_2^n}{\lVert \vect{q}_n^1 \otimes \vect{q}_2^n \rVert}$
    $\cdot \frac{1}{n_{int}}$ \Comment{increment quaternion} \\
    \small{
    \For{\forcond}
    {
        $\vect{q}_{inc}[k] = [q_{inc}^r, k \cdot{\vect{q}_{inc}^{im}}]$ \\ 
        $\vect{q}_{inc}[k]$ = $\frac{\vect{q}_{inc}[k]}{\lVert \vect{q}_{inc}[k] \rVert}$ \\ 
        $\vect{q}_{interp}[k]$ = $\frac{\vect{q}_n^1 \otimes \vect{q}_{inc}[k]}{\lVert \vect{q}_n^1 \otimes \vect{q}_{inc}[k] \rVert}$ \\ 
    }
    }
\caption{Attitude interpolation $\vect{q}_n^1$ $\to$  $\vect{q}_n^2$}
\label{Algorithm:Attitude Interpolation Algorithm}
\end{algorithm} \vspace{-0.4cm}

\newcommand{\forcondd}{$k=0$ \KwTo $(K-1)$}
\begin{algorithm}[ht]
\DontPrintSemicolon
\KwInput{true trajectory samples \small{$\{\vect{p}^n_{m}[k],\vect{\Psi}_{m}[k]\}_{k=0}^{K-1}$}}
\KwOutput{inertial increments  \small{$\{\vect{q}_t[k], \vect{q}_v[k]\}_{k=0}^{K-1}$  }}

    \small{
    \For{\forcondd}
    {
        $\vect{v}^n[k] = (\vect{p}_{m}^n[k] - \vect{p}_{m}^n[k-1]) * \frac{1}{dt}$  \\
        $\vect{q}_t[k] = \vect{h}(\vect{F}(\vect{\Psi}_{m} [k]) \cdot \vect{F}(\vect{\Psi}_{m} [k-1])^T) $ \\
        $\vect{\Delta v}^b(k) = \vect{F}(\vect{\Psi}_{m} [k]) \cdot (\vect{v}^n[k] - \vect{v}^n[k-1])$ \\
        $\vect{q}_v(k) = \vect{\Delta v}^b[k] - \vect{F}(\vect{\Psi}_{m} [k]) \cdot (0,0,g \cdot dt)^T $
    }
    }
    \caption{IMU increment generation }
\label{Algorithm:IMU increments generation}
\end{algorithm}
\subsection{Simulation Results} \label{Simulation Results}

\subsubsection{Simulation setup}  \label{Simulation setup}
We first validate our claims (Hypotheses \ref{Hypothesis_1}, \ref{Hypothesis_2}) in a simulated environment, by running 50 rollouts of the same episode on the following scenarios: 

\begin{itemize}
    \item \textbf{Noise-Less:} In this setting, we generate clean, high-frequency inertial sensor measurements, as described in section \ref{subsec: Simulating_Uncertainties}, with no added noise. We then apply our navigation filter \ref{subsec:SD algorithm Our}  algorithm in order to  perfectly reconstruct the trajectory. \label{Noise_Less_sim}
    \item \textbf{Sensor Fusion Noise:} In this setting, too, we generate a clean, high-frequency trajectory, as described for the \textit{Noise-Less} setting, but we add IC and inertial sensor errors according to \eqref{eq: att noise}.
    The IC error values are:
    $\vect{\Delta p}_{IC}=(5,5,5)$ [cm],
    $\vect{\Delta v}_{IC} =(1,1,1)$ [$\frac{cm}{sec}$],
    $\vect{\Delta \Psi}_{IC} = (4,4,5)$[deg]. 
    In addition, we set the aiding sensor noise to be 
    $\vect{\Delta p}_{err} = \vect{\Delta}(x,y,z)=(5,5,5)$ [cm] for position and
    $\vect{\Delta \Psi}_{err} = \vect{\Delta}(\phi,\theta,\psi)=(1,1,5)$ [deg],
    according to \eqref{eq: primary sensor noise in sim}, which reflects a relatively low measurement noise setting. We then apply our sensor-fusion EKF algorithm (see section \ref{subsec:SD algorithm Our}) in order to reconstruct the trajectory, which now contains localization noise. This type of simulation allows us to mimic the uncertainty in a typical INS.
    \item \textbf{Extreme Noise:} This setting is similar to \textit{Sensor Fusion Noise} but here the aiding sensor measurements contained a higher noise level, i.e.,
    $\vect{\Delta p}_{err} = \vect{\Delta}(x,y,z)=(8,8,8)$ [cm] for position and
    $\vect{\Delta \Psi}_{err} = \vect{\Delta}(\phi,\theta,\psi)=(1,1,10)$ [deg]; all the steps from the previous section (see \eqref{eq: primary sensor noise in sim}) were followed under these conditions. 
    \item \textbf{Increasing attitude noise:} We evaluated the effect of attitude noise, in particular $\Delta \psi$, where we compared the performance of $agnet_1$ and $agnet_2$ under increasing levels of measurement noise, i.e., $\vect{\Delta \Psi}_{err} = \vect{\Delta}(0,0,\psi)=(0,0,0 \rightarrow 50)$ [deg]. Each point in Figure \ref{fig:agent_1_2_simulation_results} is a mean of 50 grading episodes. The IC and other errors were the same as in section \ref{Simulation setup} above.
\end{itemize}

\subsubsection{Simulation results analysis} 
\begin{figure}
    \centering
    \subfloat[]{\includegraphics[width=0.3\linewidth]{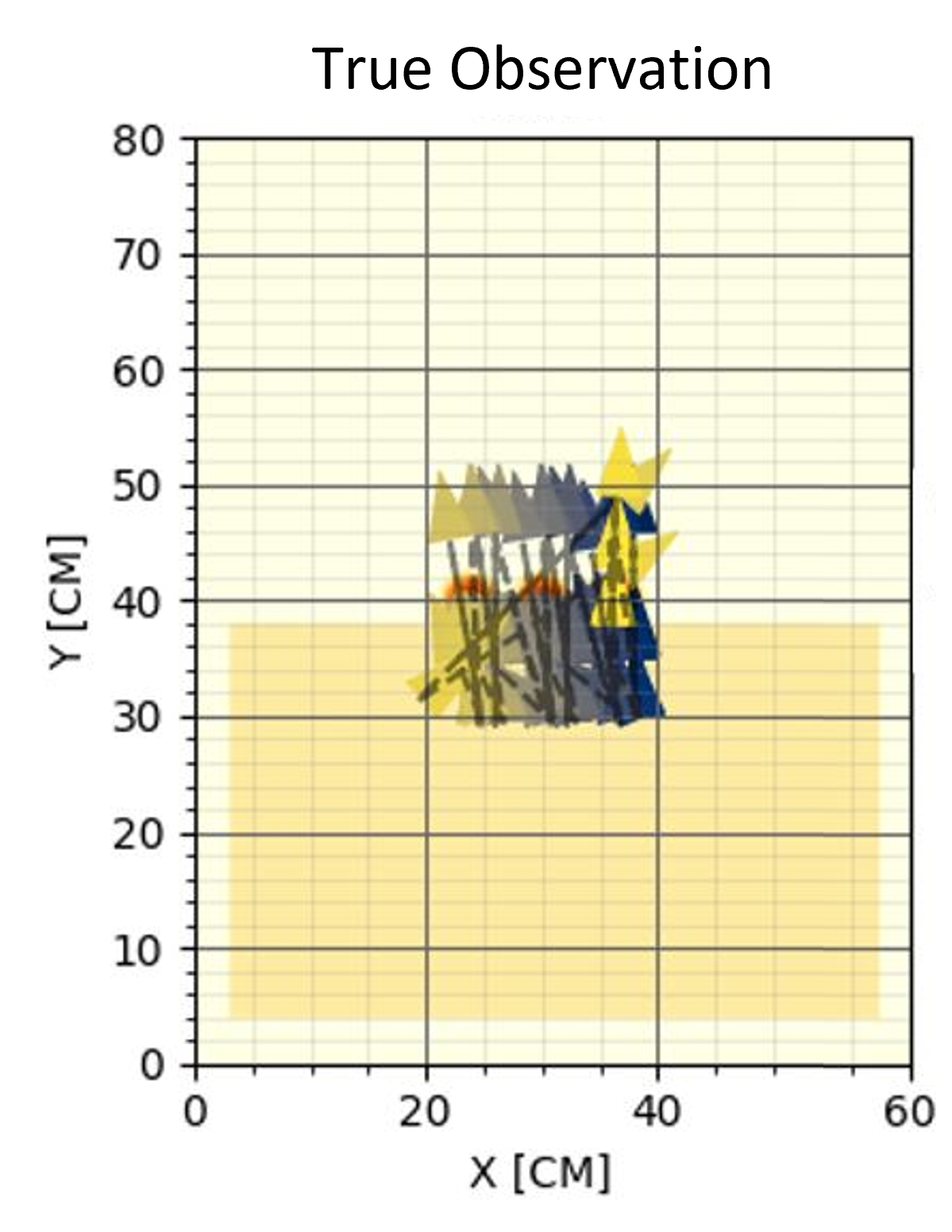} \label{fig:nav_episode_with_true_and_noise_1_2_a}} 
    \subfloat[]{\includegraphics[width=0.3\linewidth]{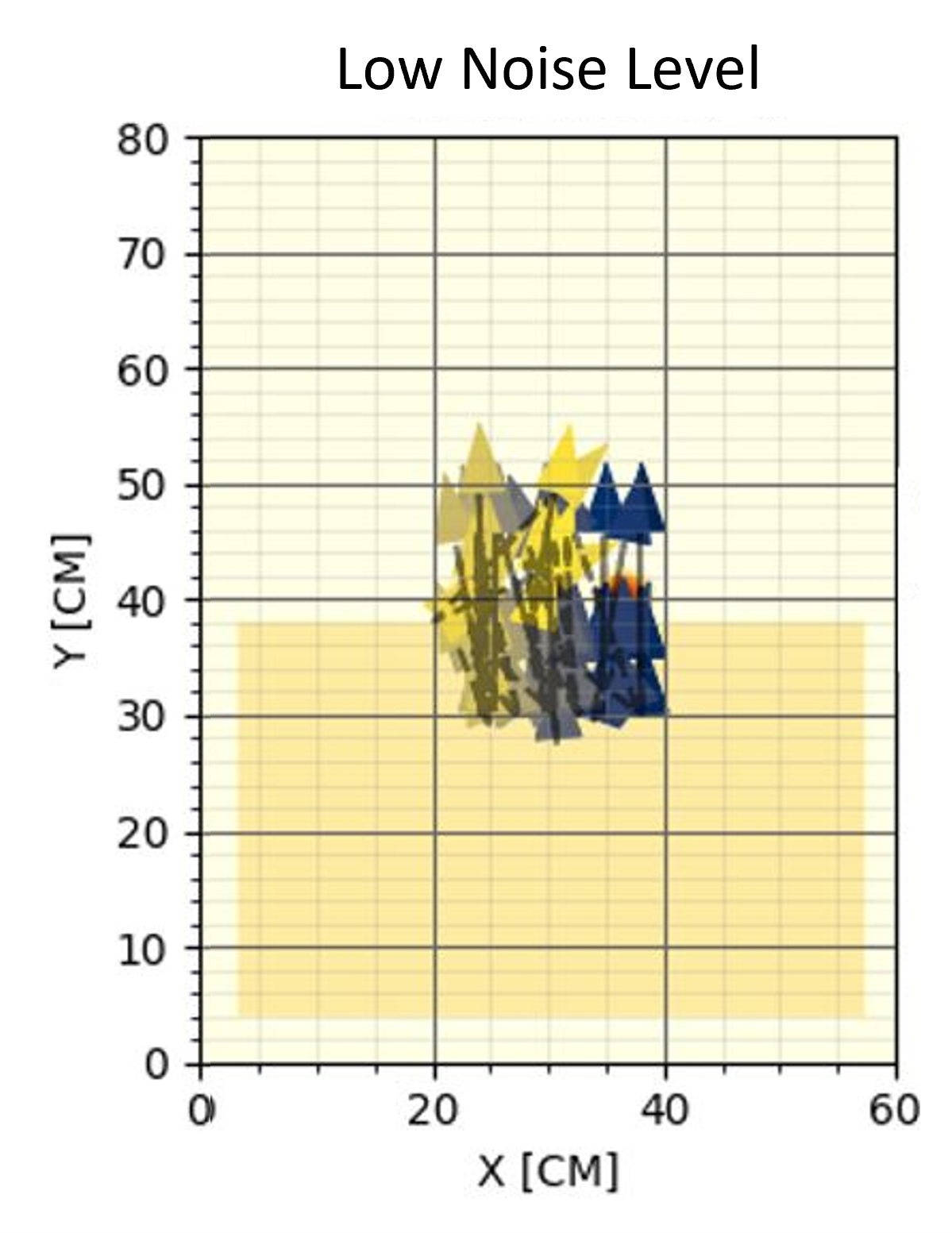} \label{fig:nav_episode_with_true_and_noise_1_2_b}}
    \subfloat[]{\includegraphics[width=0.3\linewidth]{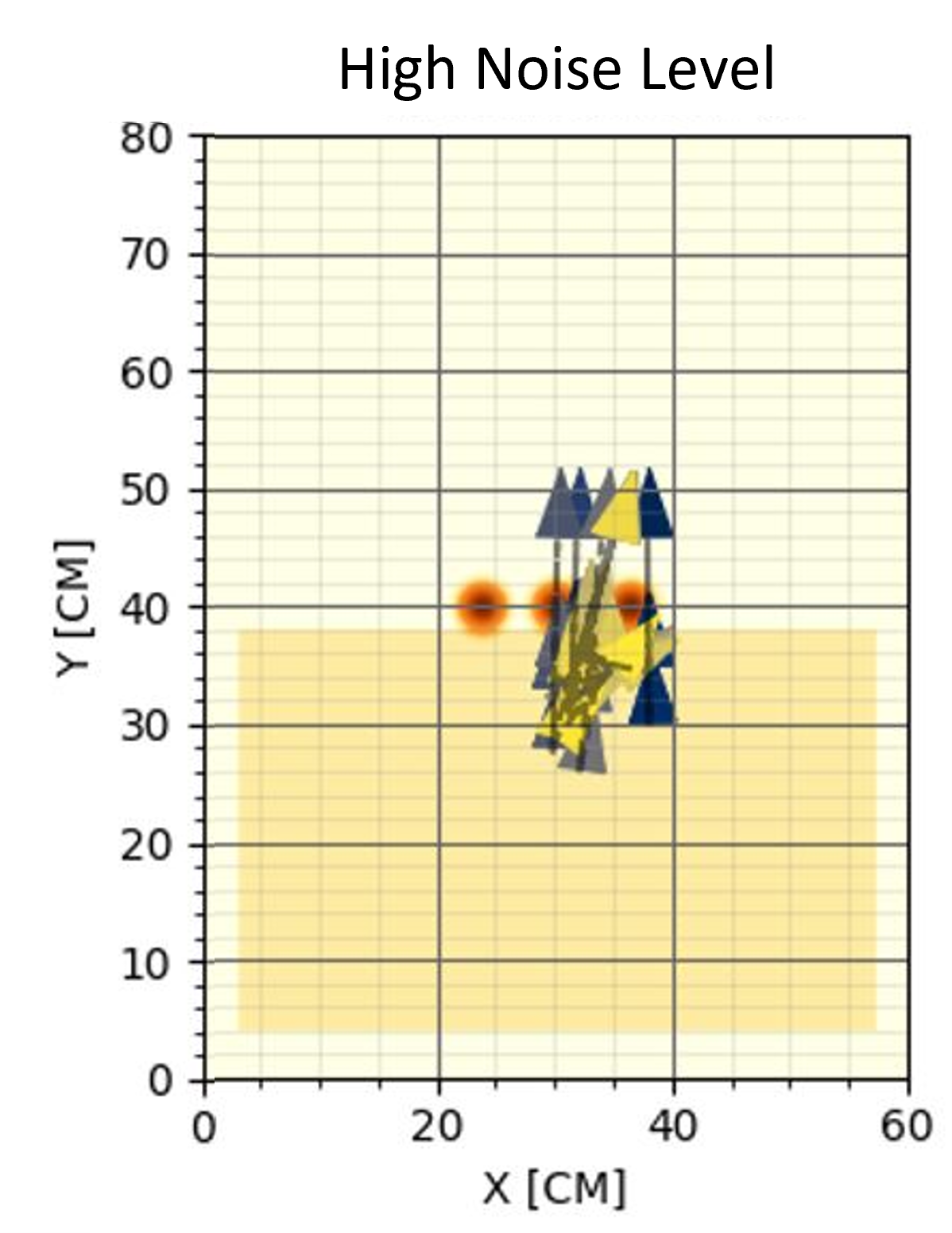} \label{fig:nav_episode_with_true_and_noise_1_2_c}} \caption{$agent_1$ simulation results with respect to the followed trajectory. The 
    number of arrows represents the number of grading legs in an episode. 
    Higher number suggests sub-optimality as it took the same agent \textit{more time} to preform the \textit{same task}.
    (\textbf{a}) Visualization of the followed trajectory without noise. (\textbf{b}) Visualization of the \textit{same} scenario with low noise level. It is noticeable that the number of actions it took the agent to complete the task is higher in the presence of noise. (\textbf{c}) Visualization of the \textit{same} scenario with high noise level. It is noticeable that the agent \textit{did not} complete the task, as some of the sand piles are left untouched. 
     }
    \label{fig:nav_episode_with_true_and_noise_1_2}
\end{figure}

\begin{figure} 
    \centering
    \subfloat[]{\includegraphics[width=0.45\linewidth]{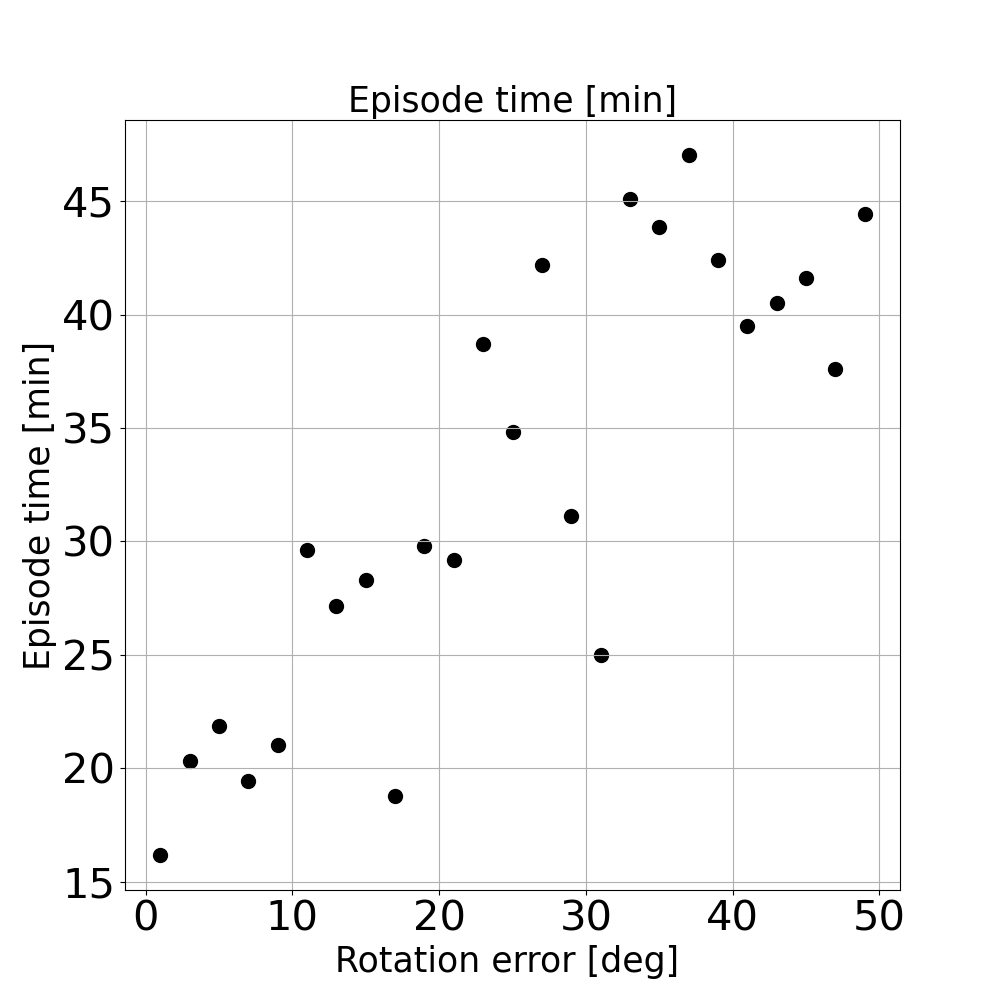} \label{fig:agent_1_simulation_results_a}} 
    \subfloat[]{\includegraphics[width=0.45\linewidth]{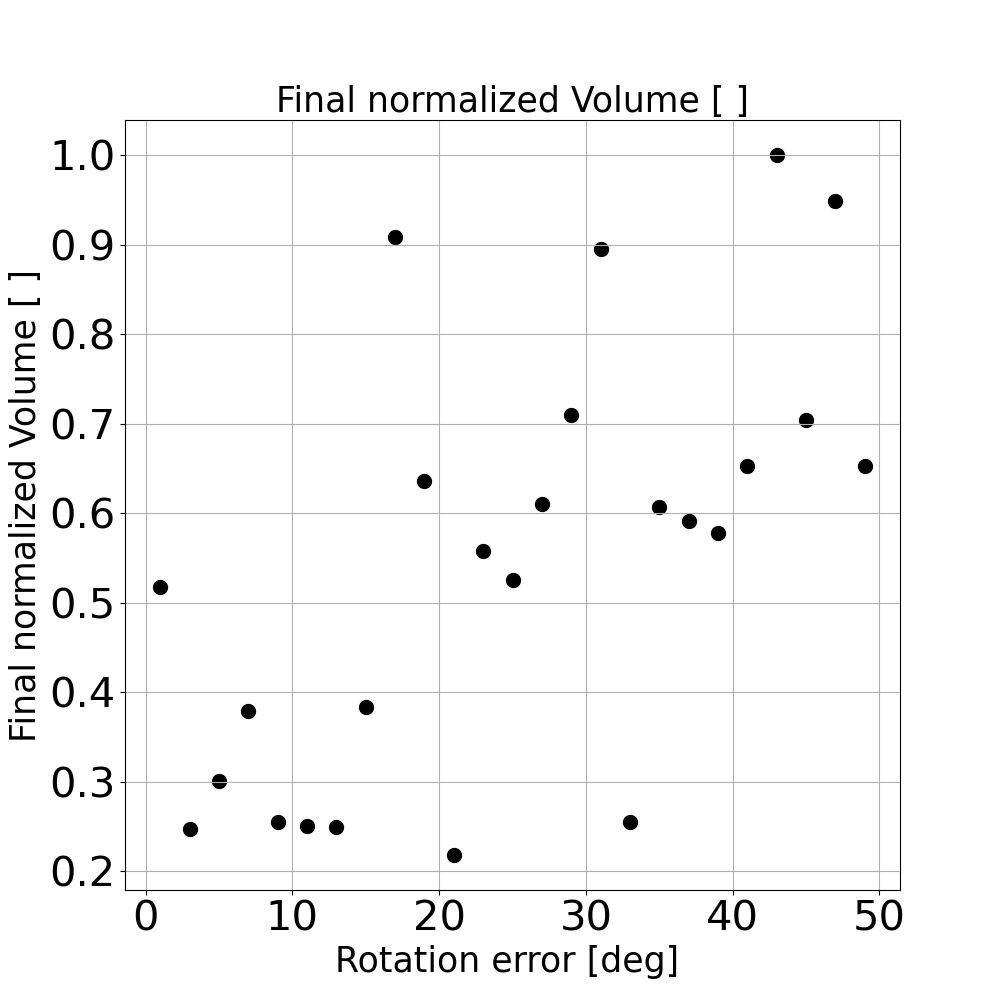} \label{fig:agent_1_simulation_results_b}} \\
    \subfloat[]{\includegraphics[width=0.45\linewidth]{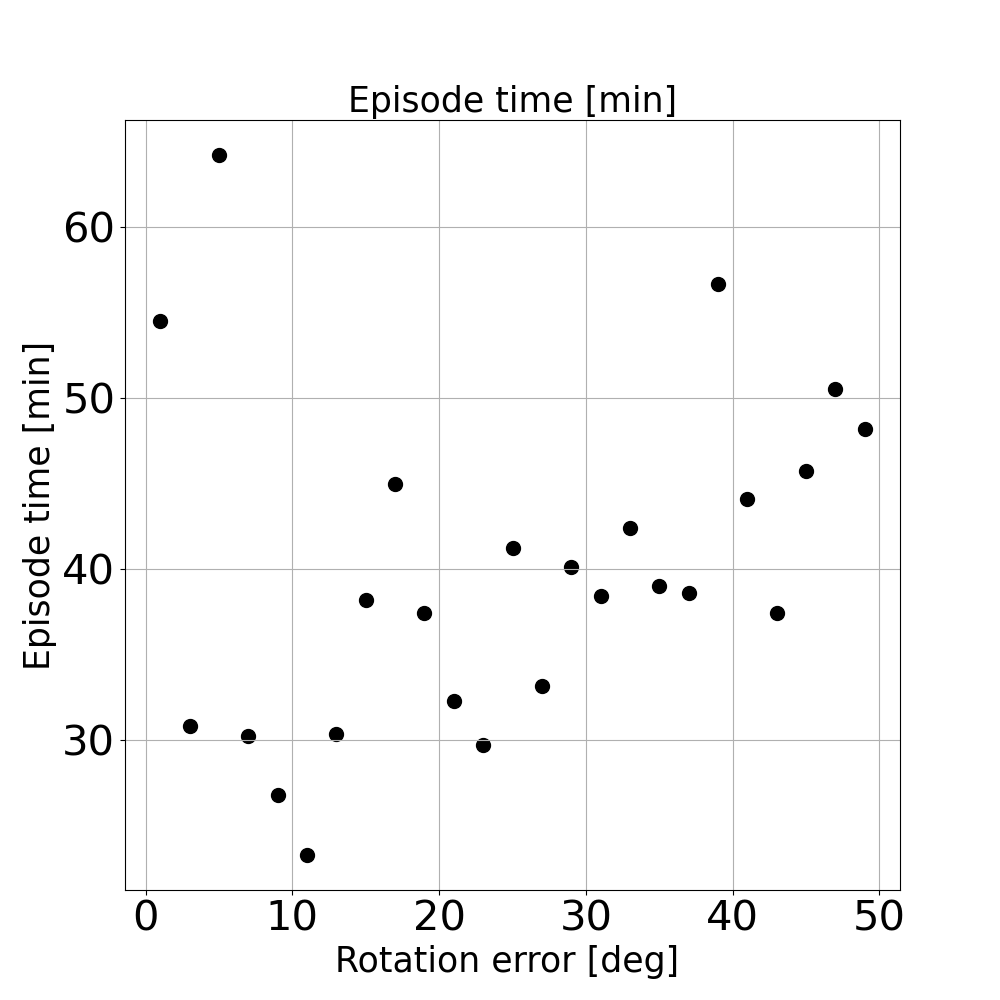} \label{fig:agent_2_simulation_results_c}}
    \subfloat[]{\includegraphics[width=0.45\linewidth]{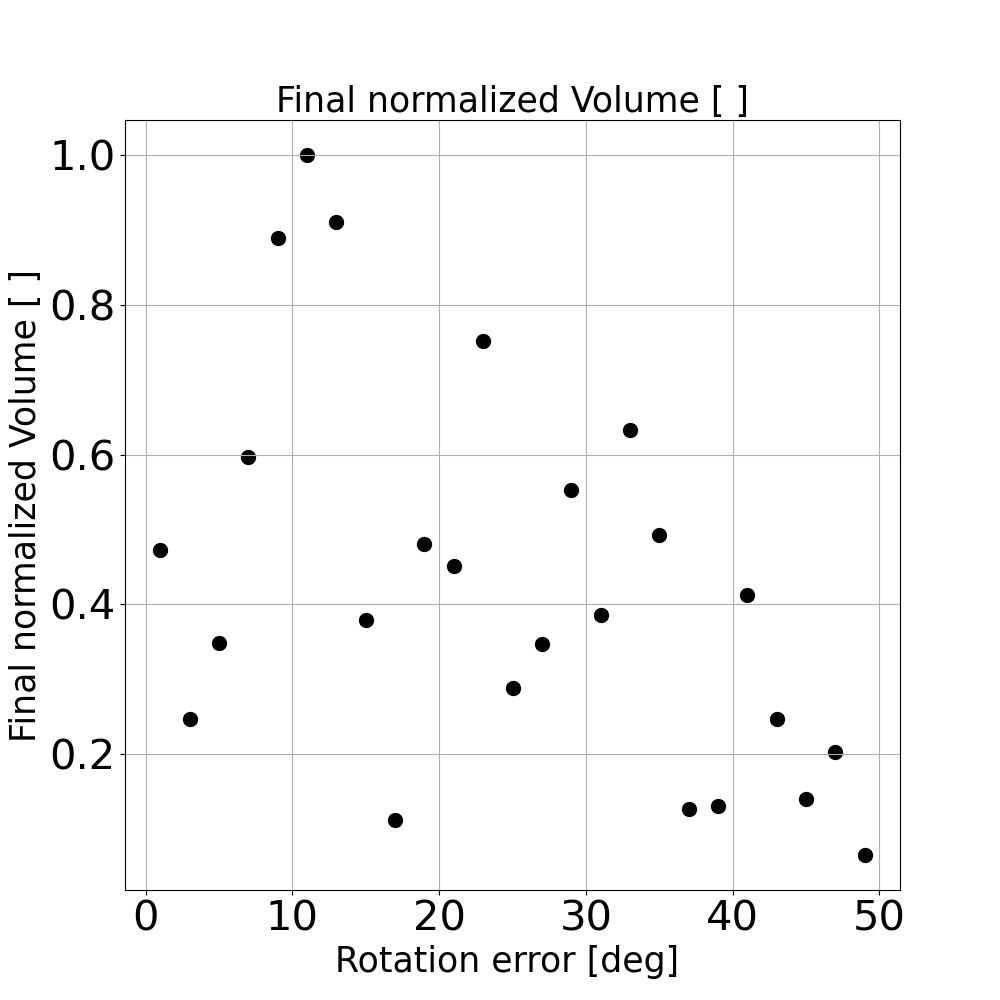} \label{fig:agent_2_simulation_results_d}}
    
    \caption{Simulation of the performance of $agent_1,agent_2$ under increasing noise levels. (\textbf{a}-\textbf{b}) When the noise level increased, so did the (\textbf{a}) time it took $agent_1$ to complete the task increased \textit{rapidly} and (\textbf{b}) the volume that remained uncleared, suggesting sub-optimal performance under uncertainties.
    (\textbf{c}-\textbf{d}) As the noise level increases, the (\textbf{c}) time it took  $agent_2$ to complete the task increased in a moderate rate, while the (\textbf{d}) total uncleared volume did not increase, meaning that $agent_2$ is more robust to noise. 
     }
    \label{fig:agent_1_2_simulation_results}
\end{figure}

\begin{itemize}
    \item \textbf{Noise-Less:} This type of simulation is intended to serve as a baseline comparison method. As such, we consider actions taken under this setting as ideal, as can be seen in Figure \ref{fig:nav_episode_with_true_and_noise_1_2_a} 
    \item \textbf{Sensor Fusion Noise:}
    The results of $agent_1$ appear in  Figure \ref{fig:nav_episode_with_true_and_noise_1_2_b}, where it took the agent, on average over the 50 rollouts, 20\% \textit{more time} to complete the same task w.r.t. the setting without noise.
    \item \textbf{Extreme Noise:} The results of $agent_1$ appear in Figure \ref{fig:nav_episode_with_true_and_noise_1_2_c}, where the agent did not complete the task, as some sand piles were left unattended.
    \item \textbf{Increasing attitude noise:} Figures \ref{fig:agent_1_simulation_results_a} and \ref{fig:agent_1_simulation_results_b} describe the behaviour of $agnet_1$. Notice that as the measurement noise level increases, both the total time to complete the episode and the total uncleared volume increase as well. This finding supports Hypothesis \ref{Hypothesis_1}. Figures \ref{fig:agent_2_simulation_results_c} and \ref{fig:agent_2_simulation_results_d} describe the behaviour of $agnet_2$. Notice that though the noise level increases, the remaining uncleared volume decreases. We contend that this reduction highly correlates to the moderate increase in time  spent on \textit{\textbf{clearing sand}} in the case of $agent_2$. However, we argue that this is not the case for $agent_1$. In other words, $agent_2$ is more efficient in the presence of noise compared to $agent_1$, which supports Hypothesis \ref{Hypothesis_2}.
\end{itemize}

    \subsection{Scaled Prototype Results} \label{Experiments}
\begin{figure}[ht]
    \centering
        \includegraphics[width=0.48\linewidth]{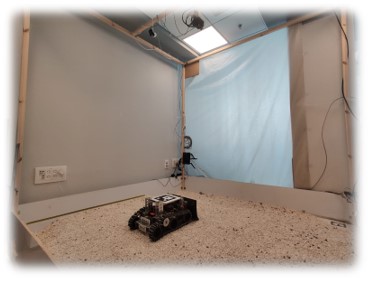}
        \includegraphics[width=0.48\linewidth]{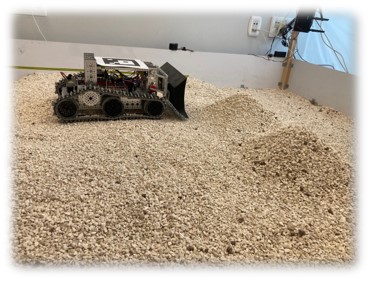}
    \caption{Our lab experimental setup. Left - image of the measurement vision system located above the sand box and our dozer robot. Right - image of the robot as it approaches the sand pile in a grading episode.}
    \label{fig: Lab_experimental_setup} 
\end{figure}

In order to validate our Hypotheses \ref{Hypothesis_1}, \ref{Hypothesis_2} under real-world conditions, we designed and built a $1:9$ scaled prototype environment (see Figure \ref{fig: Lab_experimental_setup}) that mimics several key aspects of a real-world environment. Our environment includes a RGB-D camera, mounted on top of a $250\times250cm$ sandbox and a scaled dozer prototype $60\times40cm$ in size. The RGB-D camera captures the entire scope of the target area and provides a realistic dense heightmap of it. In addition, it provides accurate pose measurements of the dozer by using an ArUco marker \cite{romerospeeded}. The dozer prototype itself is equipped with an inertial sensor VN-100 \cite{VectorNav}, which outputs the velocity and angular increments. These increments, which are processed by the INS algorithm, together with the pose output from the camera, are fed to the EKF filter, as described in Section \ref{subsec:SD algorithm Our} and illustrated in Figure \ref{fig: overall_system}. The heightmap images (observations) and fused poses from the EKF are then fed to the agent to predict its next actions. The agent then implements a low-level controller, moving according to the chosen destinations in a closed control loop manner. 
The three noise scenarios that were taken into consideration are:
(i) \textbf{Noise-Less}: In this scenario, we used the measurements of the aiding system (with its inherent noise) as is, i.e., without additional noise. 
(ii) \textbf{Sensor Fusion Noise}: Here, we added additional errors (other than the inherent noise of the aiding system measurements) at the level of
$\vect{\Delta p}_{err} = \vect{\Delta}(x,y,z)=(5,5,5)$ [cm] for position and
$\vect{\Delta \Psi}_{err} = \vect{\Delta}(\phi,\theta,\psi)=(0,0,5)$ [deg].
(iii) \textbf{Extreme Noise}: Here, too, additional errors (other than the inherent  noise) were added, but at the level of
$\vect{\Delta p}_{err} = \vect{\Delta}(x,y,z)=(5,5,5)$ [cm] for position and
$\vect{\Delta \Psi}_{err} = \vect{\Delta}(\phi,\theta,\psi)=(0,0,10)$ [deg]. 

\subsubsection{Decision-Making Under Uncertainty} \label{subsec: Decision_making Under Uncertainty}
\begin{figure}[ht] 
\centering
        \includegraphics[width=0.6\linewidth]{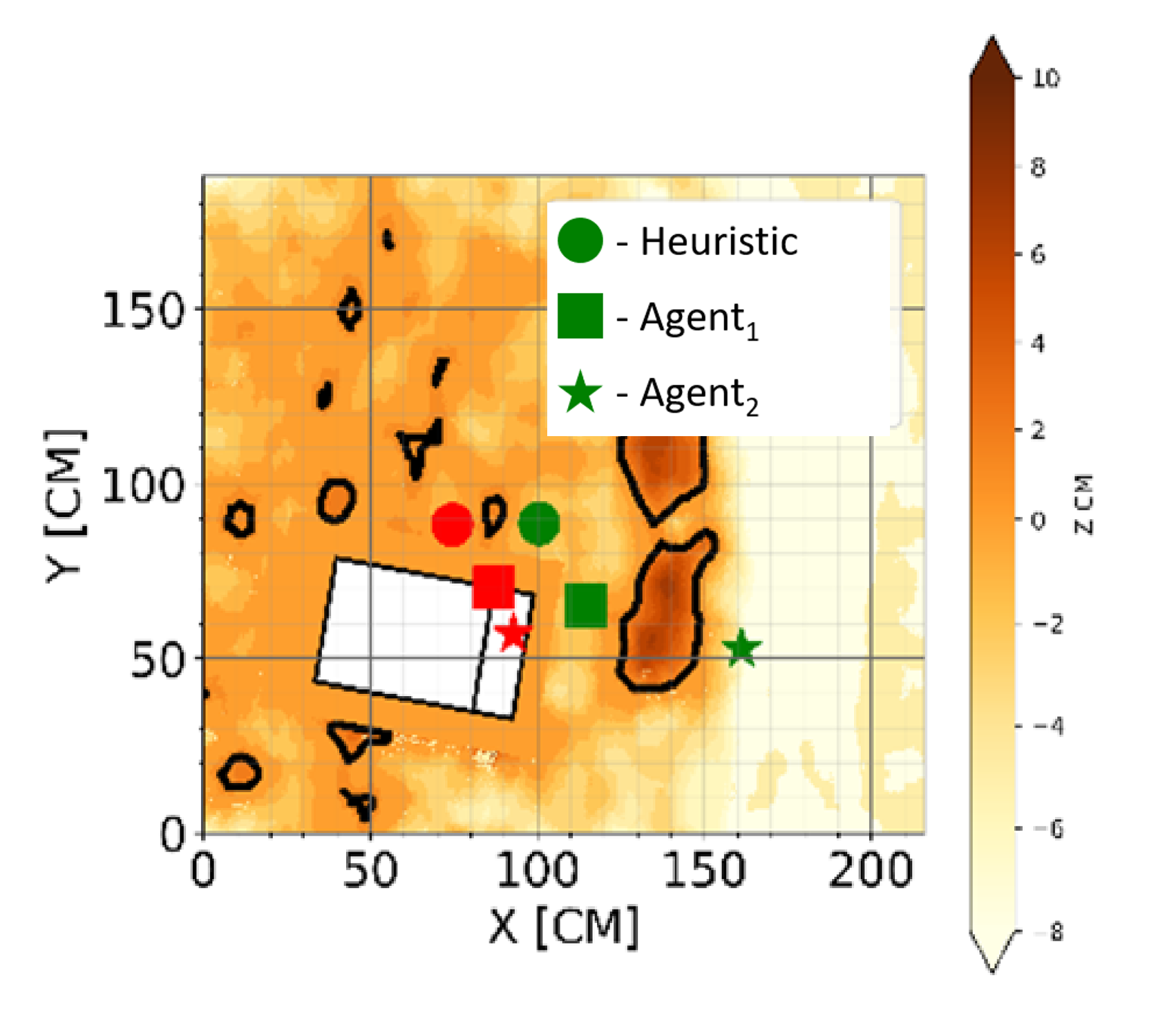}
        \caption{Here, we show an example of decisions in the form of way-points. Green markers represent the way-point an agent must reach in order to successfully grade sand for a certain leg. The red marker represents the next leg's starting way-point the agent must reach after successfully arriving at the green way-point. This figure compares the performance of three agents: (i) A heuristic agent presented in \cite{ross2021agpnet}, represented by a round marker, for comparison. (ii) $agent_1$ from \cite{miron2022towards}, represented by a square marker. (iii) $agent_2$, represented by a star marker. Here, agents (i) and (ii) made sub-optimal decisions and will, therefore, not grade any sand for this specific leg. Agent (iii) made a successful decision and will, therefore, grade sand in this leg (indicated by the green star marker). In addition, it successfully chose the next way-point (indicated by a red star marker).}
        \label{fig:decisions_real_exp} 
\end{figure}

In these experiments, we examine the quality of the decisions made by $agent_1$ and $agent_2$ under the scenarios described in \ref{subsec: Decision_making Under Uncertainty}. 
\newline \textbf{Experiment setup:} 
\newline In each scenario, 50 rollouts were collected; the results are shown in Table \ref{decision_making_table}. 
An action is classified as successful if over 50\% of the blade capacity is filled with sand, i.e., sand will be spread in this leg. Figure \ref{fig:decisions_real_exp} presents a visualization of successful and unsuccessful decisions.
\newline \noindent \textbf{Results analysis:}
\begin{itemize}
    \item \textbf{Noise-Less}: In this scenario, both agents made equally successful decisions, meaning $agent_1$ was a good selection for benchmarking and comparison. Moreover, though $agent_2$ was trained to cope with noise, its performance did not degrade due to the absence of noise (Table \ref{decision_making_table}, first row).
    \item \textbf{Sensor Fusion Noise}: In this scenario, $agent_2$ outperformed $agent_1$ by a large margin, making, on average, 40\% better decisions (Table \ref{decision_making_table}, middle row). Here, the fusion algorithm was enabled for both agents, meaning that under residual noise, $agent_1$ made many unsuccessful decisions compared to $agent_2$. From this scenario we conclude that our training method can improve decision-making under uncertainties in real scenarios.
    \item \textbf{Extreme Noise}:
    In this scenario, both agents made some successful decisions and some unsuccessful decisions. Here, too, the performance of $agent_2$ was on par with that of $agent_1$ ($\approx50\%$ for both; Table \ref{decision_making_table}, bottom row). This proves the first hypothesis, that noise reduces the performance of agents on the grading task. Moreover, we see that under extreme noise conditions, $agent_2$'s performance was degraded.

\end{itemize}


\begin{table}[ht]
\centering
\begin{tabular}{|l|l|l|}
\hline
\textbf{Accurate Decisions}                                                              & $agent_1$ & $agent_2$ \\ \hline
Noise-Less                                                     &\;\;\;\;96\%         &\;\;\;\;98\%         \\ \hline
\begin{tabular}[c]{@{}l@{}}Sensor Fusion Noise \end{tabular} & \;\;\;\;50\%       & \;\;\;\;90\%       \\ \hline
\begin{tabular}[c]{@{}l@{}}Extreme Noise \end{tabular}  & \;\;\;\;52\%       & \;\;\;\;56\%       \\ \hline
\end{tabular}

\caption{The percentage of successful decisions made by each of the two agents in three scenarios (presented as the mean values of successful decisions taken by the agents out of 50 roll-outs). As expected, without noise, both agents exhibit the same level of performance, i.e., their degree of successful decision-making was equal. In the presence of noise, though, $agent_2$ made a successful decision in 90\% of the cases, while $agent_1$ managed to do so only in 50\% of them.}

\label{decision_making_table}
\end{table} 

\subsubsection{Full Trajectory Under Uncertainty} \label{subsec: Autonomous Grading Under Uncertainty}
In these experiments, we aim to include all the aspects of a real scenario, where not only the quality of the decisions are taken into account (as in Table \ref{decision_making_table}), but also the full cycle of perception, planning and path following in the presence of noise. 
\newline \textbf{Experiment setup:} 
\newline \noindent We measured the total time it took an agent to complete the episode on the three scenarios discussed in \ref{subsec: Decision_making Under Uncertainty}. In each scenario, 3 rollouts were collected for both $agent_1$ and $agent_2$. The results are shown in Table \ref{full_trajectory_table_results}, which presents the mean time it took to complete the episode in each scenario.
\newline \noindent \textbf{Results analysis:}
\begin{itemize}
    \item \textbf{Noise-Less:} Both agents took approximately 45 seconds to complete the task (Table \ref{full_trajectory_table_results}, first row).
    \item \textbf{Sensor Fusion Noise:} The performance of $agent_2$ degraded w.r.t the scenario without noise (105 sec rather than 45 sec), since residual noise still exists even though the fusion algorithm was enabled. $agent_1$ fared much worse: after making wrong decisions at the beginning of the episode, it followed the wrong path and did not even detect the next way-points in the episode. 
    \item \textbf{Extreme Noise:} $agent_2$ completed the task after 156 seconds, highlighting the effect of noise on the path-following algorithm. As $agent_2$ was trained to make better decisions under uncertainties, it took more time to finish the task due to errors in the control loop. Here, too, $agent_1$ diverged, for the same reasons as in the \textit{Sensor Fusion Noise} scenario. Moreover, under \textit{extreme noise} settings, $agent_2$'s performance degraded the most w.r.t the two prior scenarios (156 sec vs. 105, 45 sec).
    
\end{itemize}

\noindent From these experiments, we conclude that: (i) Though $agent_2$ made successful decisions in the \textit{Extreme Noise} scenario, it still exhibits degraded performance compared to the \textit{noise-less} scenario, since noise also affects the path-following algorithm. (ii) In the \textit{Sensor Fusion Noise} and \textit{Extreme Noise} scenarios, $agent_1$ diverged and did not complete the task due to unsuccessful decisions and did not manage to recover for the rest of the episode.

    
    

\begin{center}
\begin{table}[ht]
\centering
\begin{tabular}{|l|l|l|}
\hline
\textbf{Episode Length [sec]}                                                              & $agent_1$ & $agent_2$ \\ \hline
Noise-Less                                                     & \;\;\;\;44        & \;\;\;\;45        \\ \hline
\begin{tabular}[c]{@{}l@{}}Sensor Fusion Noise \end{tabular} & diverged  & \;\;\;105       \\ \hline
\begin{tabular}[c]{@{}l@{}}Extreme Noise \end{tabular}  & diverged  & \;\;\;156       \\ \hline
\end{tabular}
\caption{Results comparing two agents in three scenarios (all results are in seconds). As expected, without noise, the performance of both agents was on par. In the presence of noise, though, $agent_2$ managed to complete the task in all three cases, while $agent_1$ diverged early on in the episode.}
\label{full_trajectory_table_results}
\end{table}
\end{center} \vspace{-0.8cm}
    \section{Conclusions} \label{Conclusion}
In this work, we compare several grading policies under localization uncertainties. We observe that an agent trained 
with complete certainty of its pose will exhibit degraded performance when presented with real-world localization uncertainty at inference, as the agent is exposed to out-of-distribution observations.
To cope with this issue, we devise a novel training regime, where the agent is presented with localization uncertainty during training. Through multiple evaluations, we show that the proposed training procedure \textit{indeed} improves the performance compared to the baseline method by 40\%. 
This improved performance is further validate via rigorous experiments, both in simulation and on a real-world scaled prototype. While the presented method is applied to the autonomous grading task, it can be applicable to any autonomous vehicle task. 
As future work, training $agent_2$ to operate under extreme noise could be considered, as the bound is currently set to these conditions.




	\bibliographystyle{IEEEtran}
	\bibliography{Dozer_uncert}

\end{document}